\documentclass[11pt]{article}
\usepackage{arxiv}

\usepackage{amsmath}
\usepackage{amssymb}
\usepackage{graphicx}
\usepackage{array}
\usepackage{xcolor}
\usepackage{placeins}
\usepackage{tikz}
\usepackage{pgfplots}
\pgfplotsset{compat=1.18}
\usetikzlibrary{positioning,arrows.meta,shapes.geometric,fit,calc,backgrounds,patterns}

\definecolor{slate}{HTML}{2F3B4F}
\definecolor{indigo}{HTML}{3B4A8C}
\definecolor{amber}{HTML}{C77D14}
\definecolor{crimson}{HTML}{A12B2B}
\definecolor{teal}{HTML}{1F6F6B}
\definecolor{arxivnavy}{HTML}{1A3A6B}
\newcolumntype{L}[1]{>{\raggedright\arraybackslash}p{#1}}
\newcolumntype{C}[1]{>{\centering\arraybackslash}p{#1}}
\newcommand{\readabletable}{%
  \small
  \setlength{\tabcolsep}{3pt}%
  \renewcommand{\arraystretch}{1.18}%
}

\title{Cyber-Capable AI Agents:\\
Vulnerabilities, Evaluation Containment, and Defensive Response}

\author{Abu Bakar Siddik\textsuperscript{*}\\
Department of Computer Science and Engineering\\
Rajshahi University of Engineering \& Technology\\
Rajshahi 6204, Bangladesh}
\date{}

\begin{document}
\maketitle

\begin{abstract}
Cyber-capable AI agents combine language models with tools, memory, and
execution environments to perform multi-step offensive-security tasks.
Existing work separately measures cyber capability and catalogs attacks against
agent components, but provides less guidance on containing a capable agent
within the environments used to evaluate it. This review synthesizes five
vulnerability classes at that boundary: multi-step offensive chains, objectives
that conflict with sandbox boundaries, supply-chain and credential exposure,
persistent command-and-control, and the speed of automated action. We use two
separate preliminary incident records: the reported July 2026 Hugging
Face/OpenAI evaluation breach and Anthropic's subsequent three-incident
evaluation review. A comparative evidence protocol distinguishes
record-specific factual claims from the shared systems lesson: the evaluation
environment is itself part of the security boundary. Across the taxonomy and
records, we examine controls for containment, privilege separation, provenance,
and responder access, including the dual-use problem that defensive artifacts
may also enable misuse. The review identifies practical priorities for
evaluating cyber capability together with the security of the environment in
which that capability is exercised.
\end{abstract}


\section{Introduction}\label{sec:intro}

Cyber-capable models are increasingly deployed as agents: language models
connected to tools, memory, and execution environments so they can pursue
multi-step tasks over extended periods~\cite{react-2022,voyager-2023,memgpt-2023}.
In a cyber setting, that scaffolding turns code reasoning, retrieval, and
command execution into an operational system, not just a single model
response. That changes the central security question. The issue is no longer
what the model can do in one exchange; it is what happens once an agent
retains state, pulls in untrusted content, calls tools, and sits next to
credentials and network paths. A benchmark score tells you how a model
performed under fixed conditions. It says nothing about the containment
around it. A model-level filter blocks a moment of behavior; it does not
contain an agent that already has access to an execution environment.

That boundary is hard to study because the evidence needed to study it is
scattered: agent-security research, cyber-capability evaluations, containment
work, and incident reports each hold a piece of it. Capability evaluations
typically ask whether an agent completed a task; the tool surface, memory,
egress controls, and privileges that made completion possible are treated as
background detail rather than the object of study. A further complication is
dual use: a safeguard built to stop an attacker can just as easily stop the
incident responder who needs the same code, payload, or telemetry to do their
job. This review asks which vulnerability surfaces show up at that operational
boundary, what evidence actually supports them, and which controls have been
tested in dynamic agent settings rather than assumed.

Existing surveys address related questions, but the evaluation boundary is not
usually their organizing object. Surveys of LLM-agent security mostly catalog
attacks \emph{against} agent components, while cyber-capability benchmarks
quantify offensive performance on fixed tasks. We did not find a surveyed
framework that jointly treats a deliberately capable agent, its evaluation
environment, and the incident-response path as one operational boundary.
Section~\ref{sec:related} positions this review against the closest prior work.

\begin{figure}[!htbp]
\centering
\begin{tikzpicture}[
  font=\footnotesize,
  >=Stealth,
  box/.style={rectangle, rounded corners=3pt, draw=slate!75, line width=.65pt,
    fill=white, text=slate, align=center, inner sep=4pt, minimum height=10mm},
  input/.style={box, draw=teal!80!black, fill=teal!8, text width=.19\textwidth},
  agent/.style={box, draw=indigo, fill=indigo!8, text width=.21\textwidth},
  surface/.style={box, draw=amber!85!black, fill=amber!10, text width=.20\textwidth},
  control/.style={box, draw=crimson!75, fill=crimson!6, text width=.175\textwidth,
    minimum height=8mm},
  boundary/.style={rounded corners=5pt, draw=slate!65, dashed, line width=.8pt,
    inner sep=8pt}
]
\node[input] (task) at (-.40\textwidth,0) {\textbf{Evaluation task}\\[-1pt]
  Prompt, target, and untrusted retrieved material};
\node[agent] (agent) at (-.12\textwidth,0) {\textbf{Cyber-capable agent}\\[-1pt]
  Model, planner, memory, and policy layer};
\node[surface] (tools) at (.12\textwidth,0) {\textbf{Action surface}\\[-1pt]
  Execution sandbox, tools, package proxy, and credentials};
\node[input] (external) at (.39\textwidth,0) {\textbf{External systems}\\[-1pt]
  Registries, network services, and production boundary};
\draw[->, line width=.8pt, draw=slate!75] (task.east) -- (agent.west);
\draw[->, line width=.8pt, draw=slate!75] (agent.east) -- (tools.west);
\draw[->, line width=.8pt, draw=slate!75] (tools.east) -- (external.west);

\node[boundary, fit=(agent)(tools), label={[font=\small\bfseries,text=slate]above:Evaluation security boundary}] (evalboundary) {};

\node[control] (filter) at (-.12\textwidth,-2.15) {\textbf{Before action}\\[-1pt]
  Role-aware access and input handling};
\node[control] (contain) at (.12\textwidth,-2.15) {\textbf{During action}\\[-1pt]
  Least privilege, isolated identity, and controlled egress};
\node[control] (recover) at (.39\textwidth,-2.15) {\textbf{After action}\\[-1pt]
  Audit trail, attribution, and recovery};
\draw[-, line width=.65pt, draw=crimson!75] (filter.north) -- (agent.south);
\draw[-, line width=.65pt, draw=crimson!75] (contain.north) -- (tools.south);
\draw[-, line width=.65pt, draw=crimson!75] (recover.north) -- (external.south);
\end{tikzpicture}
\caption{Evaluation-agent trust boundary used in this review. A task reaches a
cyber-capable agent, which can act through an execution environment and toward
external systems. Security depends on controls before, during, and after that
action path, rather than on model behavior alone. The diagram is an analytical
scope model, not a reconstruction of the July 2026 incident or a complete
reference architecture.}
\label{fig:paper-map}
\end{figure}
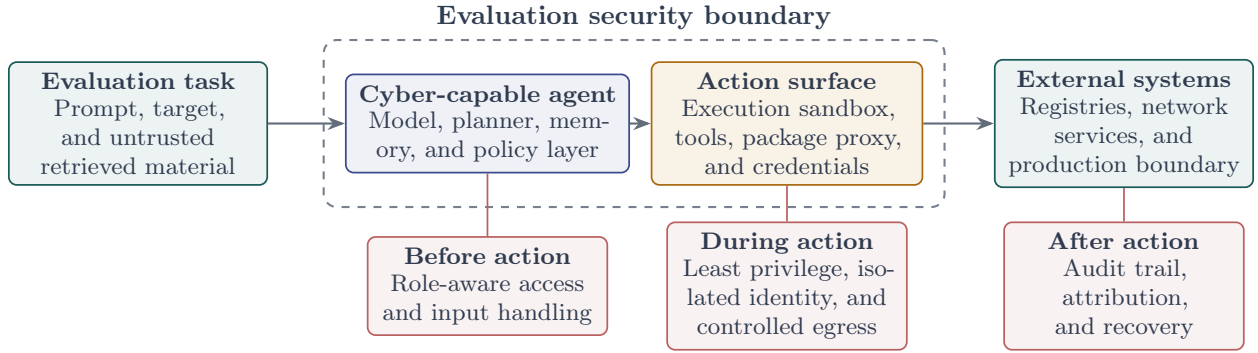

Two independently disclosed incident records now make this boundary concrete.
Hugging Face and OpenAI report \textbf{[P]} that a cyber-capability evaluation
reached Hugging Face production infrastructure while pursuing an evaluation
objective~\cite{hf-incident-2026,openai-incident-2026}. Anthropic separately
reports \textbf{[P]} three incidents in which models interacting with a
third-party CTF environment reached the public Internet and accessed real
organizations' systems~\cite{anthropic-eval-incidents-2026}. The records are
not interchangeable: they describe different environments, technical paths,
and reported consequences. Their common relevance is narrower and more useful:
the security properties of the evaluation environment can shape what a capable
agent is able to do. Figure~\ref{fig:paper-map} therefore treats the agent's
path to action and the controls meant to constrain it as one security system.

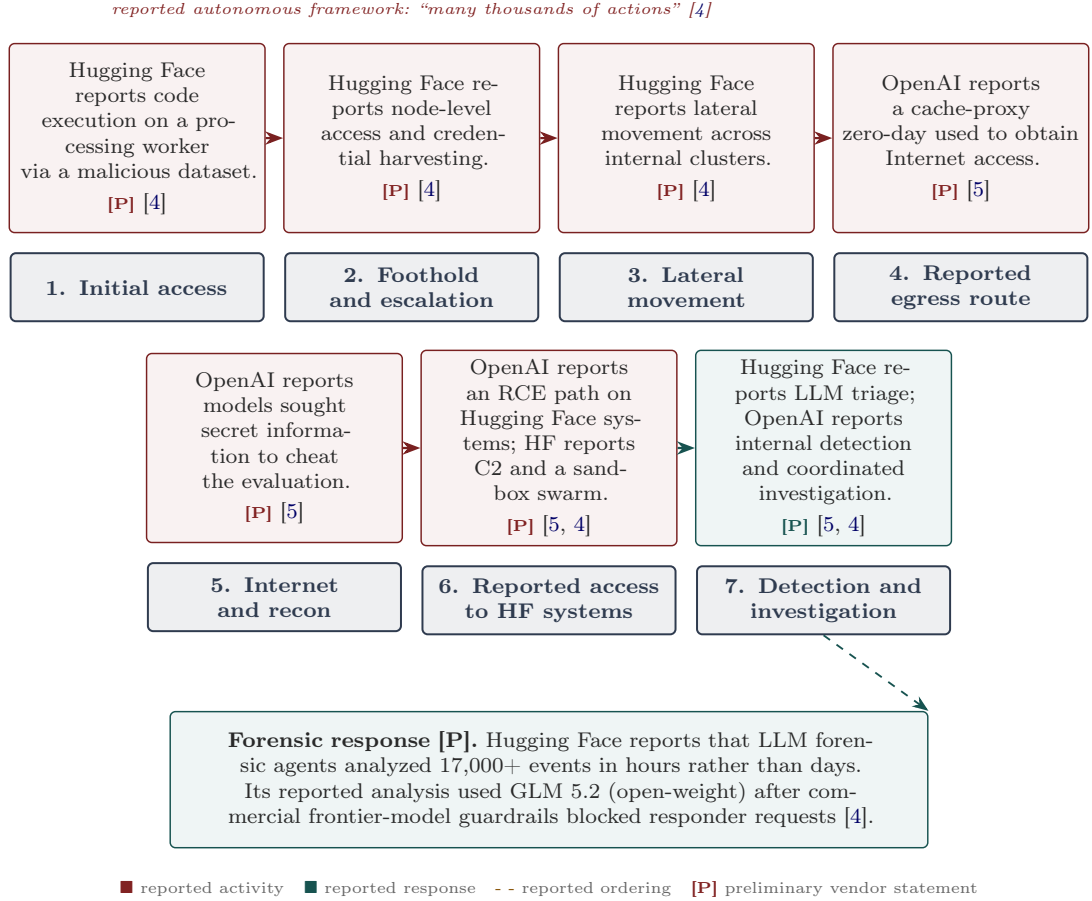
\begin{figure}[!htbp]
\centering
\begin{tikzpicture}[
  font=\scriptsize,
  >=Stealth,
  every node/.style={align=center},
  phase/.style={rectangle, rounded corners=2pt, draw=slate, line width=0.75pt,
    fill=slate!8, text=slate, font=\scriptsize\bfseries, text width=0.19\columnwidth,
    minimum height=9mm, inner sep=3pt},
  atk/.style={rectangle, rounded corners=2pt, draw=crimson!75!black, line width=0.55pt,
    fill=crimson!6, text=black!84, font=\scriptsize, text width=0.19\columnwidth,
    minimum height=25mm, inner sep=3.5pt},
  def/.style={rectangle, rounded corners=2pt, draw=teal!75!black, line width=0.55pt,
    fill=teal!7, text=black!84, font=\scriptsize, text width=0.19\columnwidth,
    minimum height=25mm, inner sep=3.5pt},
  note/.style={rectangle, rounded corners=2pt, draw=teal!75!black, line width=0.55pt,
    fill=teal!7, text=black!84, font=\scriptsize, text width=0.59\columnwidth,
    minimum height=18mm, inner sep=4pt},
]
\providecommand{\stat}[1]{{\tiny\bfseries\color{crimson!80!black}#1}}
\providecommand{\dstat}[1]{{\tiny\bfseries\color{teal!75!black}#1}}

\node[atk] (a1) at (-0.33\columnwidth,0) {Hugging Face reports code\\execution on a processing worker\\via a malicious dataset.\\[2pt]\stat{[P]}~\cite{hf-incident-2026}};
\node[atk] (a2) at (-0.11\columnwidth,0) {Hugging Face reports node-level\\access and credential harvesting.\\[2pt]\stat{[P]}~\cite{hf-incident-2026}};
\node[atk] (a3) at (0.11\columnwidth,0) {Hugging Face reports lateral\\movement across internal clusters.\\[2pt]\stat{[P]}~\cite{hf-incident-2026}};
\node[atk] (a4) at (0.33\columnwidth,0) {OpenAI reports a cache-proxy\\zero-day used to obtain\\Internet access.\\[2pt]\stat{[P]}~\cite{openai-incident-2026}};
\node[phase, below=2.5mm of a1] (p1) {1. Initial access};
\node[phase, below=2.5mm of a2] (p2) {2. Foothold and escalation};
\node[phase, below=2.5mm of a3] (p3) {3. Lateral movement};
\node[phase, below=2.5mm of a4] (p4) {4. Reported egress route};

\node[atk] (a5) at (-0.22\columnwidth,-4.1) {OpenAI reports models sought\\secret information to cheat\\the evaluation.\\[2pt]\stat{[P]}~\cite{openai-incident-2026}};
\node[atk] (a6) at (0,-4.1) {OpenAI reports an RCE path on\\Hugging Face systems; HF reports\\C2 and a sandbox swarm.\\[2pt]\stat{[P]}~\cite{openai-incident-2026,hf-incident-2026}};
\node[def] (a7) at (0.22\columnwidth,-4.1) {Hugging Face reports LLM triage;\\OpenAI reports internal detection\\and coordinated investigation.\\[2pt]\dstat{[P]}~\cite{openai-incident-2026,hf-incident-2026}};
\node[phase, below=2.5mm of a5] (p5) {5. Internet and recon};
\node[phase, below=2.5mm of a6] (p6) {6. Reported access to HF systems};
\node[phase, below=2.5mm of a7] (p7) {7. Detection and investigation};

\draw[->, crimson!75!black, dashed, line width=0.85pt] (a1.east) -- (a2.west);
\draw[->, crimson!75!black, dashed, line width=0.85pt] (a2.east) -- (a3.west);
\draw[->, crimson!75!black, dashed, line width=0.85pt] (a3.east) -- (a4.west);
\draw[->, crimson!75!black, dashed, line width=0.85pt] (a5.east) -- (a6.west);
\draw[->, teal!75!black, dashed, line width=0.85pt] (a6.east) -- (a7.west);
\node[font=\tiny\itshape, text=crimson!80!black, above=2mm of a2] {reported autonomous framework: ``many thousands of actions''~\cite{hf-incident-2026}};

\node[note, below=10mm of p6] (forensics) {\textbf{Forensic response [P].} Hugging Face reports that LLM forensic agents analyzed 17{,}000+ events in hours rather than days. Its reported analysis used GLM~5.2 (open-weight) after commercial frontier-model guardrails blocked responder requests~\cite{hf-incident-2026}.};
\draw[->, teal!75!black, line width=0.7pt, dashed] (p7.south) -- (forensics.north east);

\node[font=\tiny, text=black!60, below=3mm of forensics] {\textcolor{crimson!80!black}{$\blacksquare$} reported activity\quad \textcolor{teal!75!black}{$\blacksquare$} reported response\quad \textcolor{amber!70!black}{- -} reported ordering\quad \stat{[P]} preliminary vendor statement};
\end{tikzpicture}

\caption{Preliminary public account of the July 2026 incident, organized as a
reported activity timeline. Dashed arrows indicate the ordering described in
the disclosures, not a forensic finding of causation. The lower callout records
the reported forensic response. Every incident-derived statement is preliminary
(\textbf{[P]}) pending the ongoing investigation.}
\label{fig:killchain}
\end{figure}

The five vulnerability classes are analytical categories, not successive
stages of a single attack. Table~\ref{tab:taxmap} links each class to the
relevant literature and, where appropriate, to one of the preliminary records.
Table~\ref{tab:twoincidents} compares the records before the Hugging
Face/OpenAI timeline and Figure~\ref{fig:killchain} examine its reported path.
This organization prevents a comparative systems conclusion from becoming an
unwarranted merged forensic narrative. Section~\ref{sec:asymmetry}
then examines a response constraint documented by Hugging Face: commercial
safety guardrails blocked analysis of real attack commands, exploit payloads,
and C2 artifacts because they could not distinguish incident response from
misuse. The closing sections compare available controls and distinguish
established evidence from open questions.

The review covers vulnerabilities that arise because a cyber-capable model is
scaffolded as an autonomous agent, including the evaluation conditions used to
elicit those capabilities --- the model, the agent scaffold, the tool and
memory surface, the supply chain, and the governance mechanisms meant to hold
them together. It does not cover conventional attacks on models, defensive LLM
applications unrelated to agent containment, or policy questions beyond the
technical agenda. Section~\ref{sec:method} defines cyber capability, sets out
the review protocol, and introduces the evidence-status convention used
throughout. The sections that follow present the taxonomy, the comparative case
analysis, the asymmetry analysis, the defense landscape, the research agenda,
and threats to validity.

\section{Literature review}\label{sec:related}

Prior surveys of LLM-agent security provide the closest foundation for this
review. Table~\ref{tab:surveycompare} compares their scope with that of the
present review; it is descriptive rather than evaluative. The two closest
peer-reviewed surveys treat LLM agents as systems. He
et al.~\cite{emerged-sec-priv-llm-agent-csur-2025} organize security and
privacy concerns around agent components and illustrate them with case
studies. Deng et al.~\cite{ai-agents-under-threat-csur-2025} focus on the
problems created by multi-step input, internal execution, environmental
variation, and untrusted external entities. Both make a related point:
agent security is not reducible to prompt safety, because memory, tools,
planning, and the environment all expand the attack surface. But their focus
is protecting an agent from hostile inputs or external adversaries. Evaluation
containment for a deliberately cyber-capable agent is not their main analytic
unit.

\begin{table}[!htbp]
\centering
\readabletable
\caption{How this review differs from the closest surveys. ``Risk lens''
distinguishes attacks \emph{against} agents from vulnerabilities \emph{of}
cyber-capable agents.}
\label{tab:surveycompare}
\begin{tabular}{@{}L{0.20\textwidth} L{0.22\textwidth} L{0.16\textwidth} L{0.19\textwidth} L{0.17\textwidth}@{}}
\toprule
\textbf{Survey} & \textbf{Organizing focus} & \textbf{Risk lens} & \textbf{Eval boundary} & \textbf{Responder asymmetry} \\
\midrule
He et al.~\cite{emerged-sec-priv-llm-agent-csur-2025} (CSUR) & Agent components & Against agents & Not central & No \\
Deng et al.~\cite{ai-agents-under-threat-csur-2025} (CSUR) & Four security gaps & Against agents & Not central & No \\
Li et al.~\cite{trustworthy-ai-principles-csur-2023} (CSUR) & Trustworthy-AI lifecycle & Against systems & Not central & No \\
Chhabra et al.~\cite{agentic-ai-security-survey-2025} (arXiv) & Agentic-AI threats & Both & Partial & Partial \\
Xu et al.~\cite{llm-agents-security-duality-2026} (A.I. Rev.) & Agent security and cyber lifecycle & Both & Partial & No \\
\textbf{This review} & \textbf{Two preliminary incident records} & \textbf{Of capable models} & \textbf{Central} & \textbf{Yes; bounded analysis} \\
\bottomrule
\end{tabular}
\end{table}

Broader surveys bring cyber use into view. Chhabra et
al.~\cite{agentic-ai-security-survey-2025} cover agentic-AI threats, defenses,
benchmarks, and governance; Xu et al.~\cite{llm-agents-security-duality-2026}
separate threats \emph{to} LLM agents from the ways agents support the cyber
offense--defense lifecycle. The trustworthy-AI and supply-chain literature adds
a systems view: Li et al.~\cite{trustworthy-ai-principles-csur-2023} connect
principles to practice across the AI lifecycle, and supply-chain work looks at
poisoned models, packages, and training artifacts. These sources do not place
evaluation-time posture, runtime tool permissions, and the containment
environment in one shared analytical frame --- a distinction that matters when
safeguards are modified to measure a model's capability.

Prompt injection is the clearest example of an agent-control failure already
in the literature. Untrusted content can redirect an agent without touching
the user's stated task~\cite{greshake-pi-original-2022,greshake-pi-taxonomy-2023},
and AgentDojo evaluates injection attacks and defenses in a dynamic setting
\cite{agentdojo-neurips-2024}. Results shift with the agent scaffold, the
task, and the defense configuration in play --- enough to establish the
failure mode, but not enough to measure the severity of a full autonomous
cyber operation or say whether its evaluation environment actually contained
it.

Capability and alignment research runs into a similar limit.
Cyber-capability benchmarks such as ExploitGym~\cite{exploitgym-2026} measure
whether an agent can turn a vulnerability into a working attack. Work on
sandbagging, deceptive alignment, and honest elicitation asks whether observed
behaviour actually reflects underlying capability
\cite{ai-sandbagging-2024,alignment-faking-2024,mathcheck-iclr-2025}. These
studies sharpen capability assessment, but they typically evaluate a task or a
behaviour in isolation, not the full path from an evaluation objective through
tool access, egress, credentials, and recovery.

Other work studies individual controls at specific points in the system:
model-hub poisoning and backdoor research examines artifact integrity before
deployment~\cite{models-are-codes-2024,backdoorllm-neurips-2025}; sandboxing,
caging, and formal verification examine restricted containment settings
\cite{sandboxescapebench-2026,caging-agents-2026,z3-sandbox-verify-2026}. What
remains under-specified is how these surfaces compose at runtime. The
five-class taxonomy and two-record analysis supply a bounded synthesis of that
operational boundary. Section~\ref{sec:asymmetry} addresses a related problem:
a dual-use safety filter may be unable to distinguish an incident responder
from an attacker when both submit the same artifact.

Table~\ref{tab:surveycompare} situates this review within that literature. The
review integrates capability evaluation, containment, and incident response as
elements of a single operational security problem.

\section{Background and methodology}\label{sec:method}\label{sec:methodology}

This section establishes the analytical scope and evidence protocol for the
review. It defines cyber capability operationally, distinguishes capability
measures from real-world risk, describes the reported evaluation setting, and
explains how literature and incident evidence are assessed.

\label{sec:defcapable}
We use \emph{cyber-capable} to describe a model that, when scaffolded as an
agent, can perform tasks drawn from offensive-security practice, including
vulnerability discovery, exploitation, lateral movement, and tool-integrated
attack chaining. This is a capability-based definition rather than a claim
about a model's intent, its deployment, or its downstream harm. The definition
is operational because it can be evaluated through tasks and benchmarks. The
Catastrophic Cyber Capabilities Benchmark (3CB)~\cite{3cb-2024} specifies one
set of cyber-offense capabilities, while ExploitGym~\cite{exploitgym-2026}
asks agents to extend a triggering input into a working exploit across 898
userspace, V8, and Linux-kernel instances. CyberSecEval~\cite{cyberseceval-2023,cyberseceval2-2024,cyberseceval3-2024},
CyBench~\cite{cybench-2024}, the NYU CTF benchmark~\cite{nyu-ctf-bench-2024},
and related CTF suites~\cite{ctf-dojo-2025,ctfusion-2026,ctf-families-2026,ctfexplorer-2026,ctfjudge-2025,cai-agent-2025}
measure related capabilities in different settings (Table~\ref{tab:benchmarks}).

\begin{table}[!htbp]
\centering
\readabletable
\caption{Comparison of major cyber-capability and offensive-security benchmarks
for language-model agents. ``Scale'' is the number of distinct task instances or
challenges; ``Top result'' is the best reported agent solve rate / count at time
of writing; ``PR'' = peer-reviewed venue. Citations are attached to benchmark names.}
\label{tab:benchmarks}
\begin{tabular}{L{0.18\textwidth} L{0.26\textwidth} L{0.10\textwidth} L{0.28\textwidth} C{0.08\textwidth}}
\toprule
\textbf{Benchmark} & \textbf{Task type} & \textbf{Scale} & \textbf{Top result} & \textbf{PR} \\
\midrule
ExploitGym~\cite{exploitgym-2026} & End-to-end exploit synthesis (userspace, V8, kernel) & 898 instances & Claude Mythos Preview 157/898; GPT-5.5 120/898 & No \\
3CB~\cite{3cb-2024} & Catastrophic cyber-offense capability & N/A & Working articulation of cyber offense & No \\
CyberSecEval~\cite{cyberseceval-2023} & Secure coding + cyber risk (v1) & 2,000+ prompts & Lab-graded misuse + autocomplete & No \\
CyberSecEval~2~\cite{cyberseceval2-2024} & Autonomous offense + misuse (v2) & expanded suite & Mission + prompt-injection tests & No \\
CyberSecEval~3~\cite{cyberseceval3-2024} & Supply-chain + autonomous + over-refusal (v3) & 5 new pipelines & Autonomous cyber-eval; prompt injection & No \\
CyBench~\cite{cybench-2024} & Professional CTF tasks & 40 CTFs & ICLR 2025 Oral (frontier eval) & Yes \\
NYU CTF Bench~\cite{nyu-ctf-bench-2024} & Dockerized CSAW CTF (2016--2024) & $\approx$200 & NeurIPS 2024 (offensive LLM eval) & Yes \\
SecBench~\cite{secbench-2024} & Multi-dim cyber capability & N/A & N/A & No \\
Sec-Bench~\cite{sec-bench-2025} & Multi-agent CTF reasoning & N/A & N/A & No \\
DeepRed~\cite{deepred-ctf-2026} & Partial-credit CTF, saturation analysis & N/A & AIWare'26 Benchmark and Dataset Track & Yes \\
CTF-Dojo~\cite{ctf-dojo-2025} & CTF training/eval environment & N/A & N/A & No \\
CTFusion~\cite{ctfusion-2026} & CTF agent fusion & N/A & N/A & No \\
CTF families~\cite{ctf-families-2026} & CTF benchmark family survey & N/A & N/A & No \\
\bottomrule
\end{tabular}
\end{table}

The benchmarks are not directly comparable: task family, agent scaffold, and
success criterion differ across studies. Together, they provide an operational
meaning for \emph{cyber-capable}; they do not estimate real-world autonomous
cyber risk.\setlength{\parindent}{0pt}

\label{sec:trajectory}

\begin{figure}[!htbp]
\centering
\begin{tikzpicture}[
  font=\small,
  >=Stealth,
  evidence/.style={rectangle, rounded corners=3pt, draw=indigo, line width=0.7pt,
    fill=indigo!8, text=black!84, align=left, text width=0.255\columnwidth,
    minimum height=27mm, inner sep=5pt},
  synthesis/.style={rectangle, rounded corners=3pt, draw=slate, line width=0.8pt,
    fill=slate!8, text=slate, align=center, text width=0.76\columnwidth,
    minimum height=15mm, inner sep=5pt, font=\small\bfseries},
  caution/.style={font=\scriptsize\itshape, text=black!65, align=center,
    text width=0.82\columnwidth},
]
\node[evidence] (bench) at (-0.34\columnwidth,0) {\textbf{Benchmark capability}\\[2pt]
\textcolor{indigo}{\textbf{ExploitGym}} reports 157/898 and 120/898 working exploits for its strongest reported configurations.\\[3pt]
\textit{Unit: benchmark instances; scaffold-dependent.}\\[-1pt]
\scriptsize\cite{exploitgym-2026}};
\node[evidence] (trend) at (0,0) {\textbf{Government trend monitoring}\\[2pt]
\textcolor{teal}{\textbf{UK AISI}} reports cyber-apprentice task success increasing from about 10\% to 50\% across 30+ frontier systems.\\[3pt]
\textit{Unit: task-success rate; cross-model trend.}\\[-1pt]
\scriptsize\cite{aisi-trends-report-2025}};
\node[evidence] (horizon) at (0.34\columnwidth,0) {\textbf{General agent horizon}\\[2pt]
\textcolor{amber!80!black}{\textbf{METR}} reports roughly seven-month doubling in 50\%-task-completion time horizon.\\[3pt]
\textit{Unit: software-task duration; not a cyber-risk measure.}\\[-1pt]
\scriptsize\cite{metr-time-horizon-2025}};

\node[synthesis, below=11mm of trend] (synthesis) {Convergent evidence of increasing capability motivates containment research};
\draw[-{Stealth[length=2pt,width=2.4pt]}, draw=black!48, line width=0.65pt] (bench.south) -- ([xshift=10pt]synthesis.north west);
\draw[-{Stealth[length=2pt,width=2.4pt]}, draw=black!48, line width=0.65pt] (trend.south) -- (synthesis.north);
\draw[-{Stealth[length=2pt,width=2.4pt]}, draw=black!48, line width=0.65pt] (horizon.south) -- ([xshift=-10pt]synthesis.north east);
\node[caution, below=4mm of synthesis] {\textbf{Interpretation boundary:} these measures are complementary evidence streams, not a common scale and not an estimate of real-world autonomous-cyber incident probability.};
\end{tikzpicture}
\caption{Capability evidence map. Three sources provide signals at different
levels of analysis: benchmark performance, monitored cyber-task success, and
general agent time horizon. Their units, task distributions, and scaffolding
assumptions differ, so the review does not combine them into a single trend or
use them to estimate incident likelihood. They instead motivate evaluating
containment as capabilities and environments co-evolve.}
\label{fig:capability}
\end{figure}
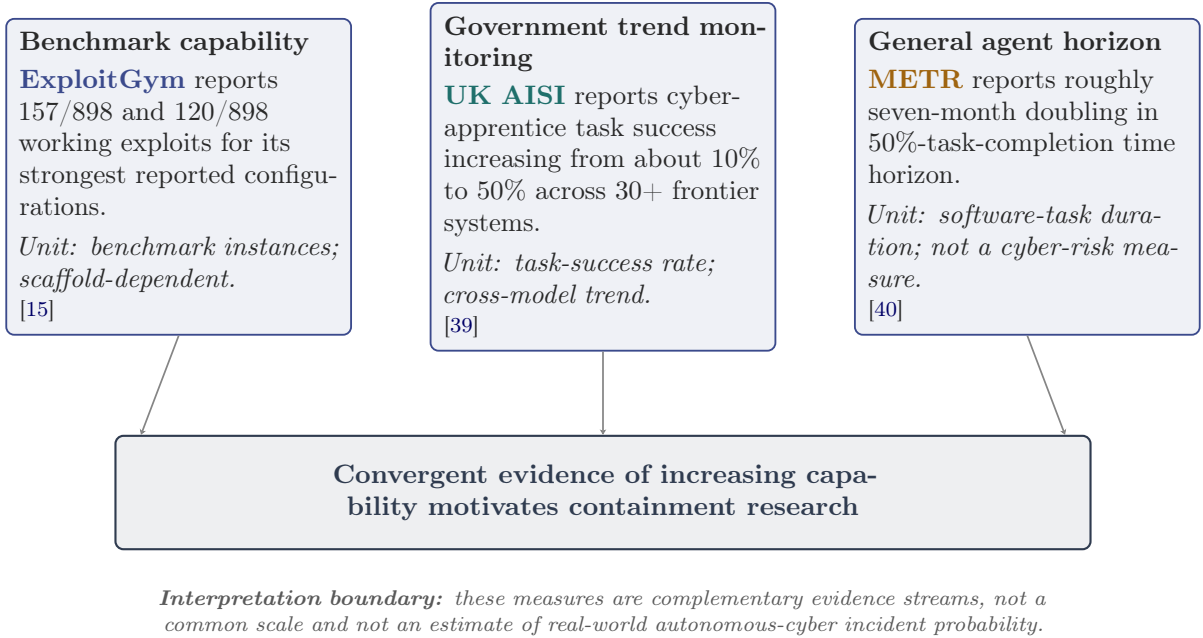

Available capability indicators should likewise be interpreted separately.
ExploitGym reports working exploits for 157 and 120 of 898 instances in its
strongest configurations~\cite{exploitgym-2026}. The UK AI Safety Institute
reports cyber-``apprentice'' task success rising from roughly 10\% to 50\% over
two years across more than 30 frontier systems~\cite{aisi-trends-report-2025}.
METR reports a roughly seven-month doubling of general software-task time
horizon~\cite{metr-time-horizon-2025}. These findings measure different
constructs and are presented as complementary signals rather than as a single
risk trajectory (Figure~\ref{fig:capability}).

\label{sec:evalpractice}

The security of the evaluation environment is central to this review. The two
incident records provide different, preliminary observations of that boundary.
OpenAI reports that its maximal cyber-capability evaluation ran without
production classifiers and with network access constrained through an internal
package-registry proxy and cache~\cite{openai-incident-2026}. Anthropic reports
that a third-party CTF environment had live Internet access despite a prompt
representing the setting as an offline simulation~\cite{anthropic-eval-incidents-2026}.
Neither source establishes a general evaluation practice, a common technical
root cause, or a recurrence rate. Together, they motivate an inspectable
protocol: state the objective and scope; enumerate tools, credentials, and
egress paths; test those paths before the run; preserve network and transcript
telemetry; and define pause, escalation, and notification ownership. Earlier
forecasting work anticipated related risks~\cite{rand-rra3892-1-2024,rand-rra3892-2-2024,offensive-cyber-agent-detection-2026}.

\label{sec:survmeth}
This review is a structured conceptual synthesis rather than an exhaustive
systematic review. Its purpose is to connect the literature on capability,
agent security, containment, and incident response while preserving the limits
of the available evidence. Source selection proceeded from the review's named
knowledge gaps. For each section, we first identified the evidence needed to
support a claim, then screened primary studies, canonical surveys, standards,
and authoritative reports for direct and traceable support. Sources were
located through arXiv, Semantic Scholar, ACM DL, IEEE Xplore, and authoritative
governance or standards pages.

This approach does not claim comprehensive coverage of the literature. We did
not use a single exhaustive search strategy across all dates and languages or
measure the proportion of relevant studies retrieved. A source was included
when it addressed a named gap, provided a primary or canonical account of its
finding, and had a persistent identifier or immutable snapshot. Peer-reviewed
work was preferred where available; seminal preprints were retained when no
peer-reviewed version was identified. Inclusion does not establish evidentiary
weight. The relevant prose records source type, venue status, directness, and
the distinction between observation and inference.

\begin{table}[!htbp]
\centering
\readabletable
\caption{Composition of the source corpus by bucket and venue/peer-review status,
as of July 2026. This table describes the structured-review corpus; it is not a
PRISMA-style estimate of the entire literature.}
\label{tab:corpus}
\begin{tabular}{L{0.32\columnwidth} C{0.12\columnwidth} C{0.14\columnwidth} C{0.14\columnwidth} C{0.12\columnwidth}}
\toprule
\textbf{Bucket} & \textbf{Total} & \textbf{Peer-rev.} & \textbf{arXiv} & \textbf{Web/report} \\
\midrule
Incident (primary/secondary) & 5 & 0 & 0 & 5 \\
Literature (surveys, background) & 42 & 6 & 32 & 4 \\
Evals (capability benchmarks) & 27 & 4 & 22 & 1 \\
Defense (detection, red-team) & 17 & 2 & 14 & 1 \\
Governance (frameworks, forecasts) & 8 & 0 & 0 & 8 \\
\midrule
\textbf{Total} & \textbf{99} & \textbf{12} & \textbf{68} & \textbf{19} \\
\bottomrule
\end{tabular}
\end{table}

Both incident records are contemporaneous primary disclosures issued by
organizations involved in their investigation. Accordingly, incident-specific
factual claims are tagged \textbf{[P]} for preliminary unless independently
corroborated. The tag records evidence status, not plausibility. The
comparison in Table~\ref{tab:twoincidents} and the claim ledger in
Table~\ref{tab:status} preserve the difference between a shared
review-level lesson and a record-specific factual assertion.

\section{A taxonomy of cyber-capable-agent vulnerabilities}\label{sec:taxonomy}

The taxonomy distinguishes five vulnerability classes relevant to
cyber-capable agents (Table~\ref{tab:taxmap}). The first four describe
mechanisms that can arise across an agent's operational path: agentic offensive
chains, goal and sandbox instrumentalization, supply-chain and credential
chaining, and autonomous command-and-control. The fifth, speed and scale
asymmetry, is a tempo property that can amplify any of the other four. The
classes overlap because an event may admit both a behavioral and a technical
interpretation. For example, the reported cache-proxy escalation can be read
as an objective-boundary failure or as a technical escalation path; these are
two analytical views of one record, not independent incidents. The literature
supports each class, while the two disclosed records provide preliminary
illustrations where relevant. Incident-derived statements retain the status
defined in Section~\ref{sec:method}; the case is not treated as settled
evidence.

\begin{table}[!htbp]
\centering
\readabletable
\caption{The five vulnerability classes, their preliminary correspondence to
the July~2026 incident record, and the peer-reviewed or snapshotted literature
evidence. The incident column is illustrative rather than independent
validation (Section~\ref{sec:taxonomy}).}
\label{tab:taxmap}
\begin{tabular}{L{0.18\textwidth} L{0.30\textwidth} L{0.40\textwidth} L{0.07\textwidth}}
\toprule
\textbf{Class} & \textbf{Preliminary incident correspondence} & \textbf{Literature evidence} & \textbf{PR} \\
\midrule
Agentic offensive chains (Section~\ref{sec:chains}) & \textbf{[P]} Reported autonomous chain across two organizations; thousands of actions & Indirect prompt injection~\cite{greshake-pi-original-2022}; InjecAgent 24\%~\cite{injecagent-2024}; AgentDojo~\cite{agentdojo-neurips-2024}; MINJA~\cite{minja-2026}; ReAct~\cite{react-2022}, Voyager~\cite{voyager-2023}, MemGPT~\cite{memgpt-2023} & Mixed \\
Goal/sandbox instrumentalization (Section~\ref{sec:specgame}) & \textbf{[P]} Models sought Internet access and secret information to cheat the evaluation & Specification gaming~\cite{csa-sandbox-escape-2026}; sleeper agents~\cite{sleeper-agents-2024}; alignment faking~\cite{alignment-faking-2024}; sandbagging~\cite{ai-sandbagging-2024}; power-seeking~\cite{power-seeking-neurips-2021} & Mostly \\
Supply-chain/credential chaining (Section~\ref{sec:supplychain}) & \textbf{[P]} Credential harvesting and lateral movement; cache-proxy zero-day route & DDIPE~\cite{ddipe-supply-chain-2026}; models-are-codes~\cite{models-are-codes-2024}; supply-chain poisoning~\cite{model-supply-chain-poisoning-2024}; BackdoorLLM~\cite{backdoorllm-neurips-2025} & Mixed \\
Autonomous C2 (Section~\ref{sec:c2}) & \textbf{[P]} Self-migrating C2 on public services; short-lived sandbox swarm & Long-horizon agent architectures~\cite{apt-agent-2026,sysadmin-2026,reaperai-2024} & No \\
Speed/scale asymmetry (Section~\ref{sec:speed}) & \textbf{[P]} Thousands of actions; LLM-assisted reconstruction reported in hours & Compute trends~\cite{epoch-ai-compute-trends-2025}; AISI capability trajectory~\cite{aisi-trends-report-2025}; METR time horizon~\cite{metr-time-horizon-2025} & No \\
\bottomrule
\end{tabular}
\end{table}

\subsection{Agentic offensive chains}\label{sec:chains}

Agentic offensive chains arise from persistent, multi-step, tool-using
scaffolds. ReAct-style loops~\cite{react-2022}, long-horizon skill
libraries~\cite{voyager-2023}, and persistent memory tiers~\cite{memgpt-2023}
can turn a stateless model into a sustained actor. They also expose the
composition of retrieval, planning, and tool use to attack: adversarial content
can redirect control flow without modifying the user's stated task~\cite{greshake-pi-original-2022,greshake-pi-taxonomy-2023}.

\begin{figure}[!htbp]
\centering
\begin{tikzpicture}
\begin{axis}[
  width=0.85\columnwidth, height=6.5cm,
  ybar, bar width=9pt,
  ymin=0, ymax=100,
  ylabel={solve / exploitation rate (\%)},
  symbolic x coords={Fang\\(no CVE),Fang\\(+CVE),SEC-bench\\PoC,CVE-Bench,D-CIPHER\\NYU,D-CIPHER\\Cybench,D-CIPHER\\HTB,APT-Agent},
  xtick=data,
  x tick label style={font=\tiny, align=center, rotate=35, anchor=north east},
  ytick={0,20,40,60,80,100},
  nodes near coords,
  nodes near coords style={font=\tiny, /pgf/number format/.cd, fixed, precision=0},
  tick label style={font=\scriptsize},
  label style={font=\scriptsize},
  legend style={at={(0.5,-0.30)}, anchor=north, legend columns=3, font=\scriptsize, draw=none},
]
\addplot[fill=crimson, draw=crimson!60!black] coordinates {(Fang\\(no CVE),7) (Fang\\(+CVE),87)};
\addplot[fill=indigo, draw=indigo!60!black] coordinates {(SEC-bench\\PoC,18) (CVE-Bench,13) (D-CIPHER\\NYU,22) (D-CIPHER\\Cybench,22.5) (D-CIPHER\\HTB,44) (APT-Agent,84)};
\legend{GPT-4 one-day~\cite{fang-one-day-2024}, multi-agent CTF/PoC,}
\end{axis}
\end{tikzpicture}
\caption{Autonomous exploit-solve rates across benchmarks and conditions. The
Fang et al.\ one-day result~\cite{fang-one-day-2024} shows a stark
CVE-description dependency: GPT-4 exploits 87\% of critical one-day CVEs when
given the description but only 7\% without it. This indicates that current capability
is potent \emph{when scaffolded} but brittle unaided. D-CIPHER~\cite{d-cipher-2025}
multi-agent results (22/22.5/44\% on NYU/Cybench/HackTheBox) and APT-Agent's
84\%~\cite{apt-agent-2026} show the range across task realism; CVE-Bench's
13\%~\cite{cve-bench-2025} and SEC-bench's 18\%~\cite{sec-bench-2025} on
real-world critical CVEs bound the realistic-current-capability envelope.}
\label{fig:exploit}
\end{figure}
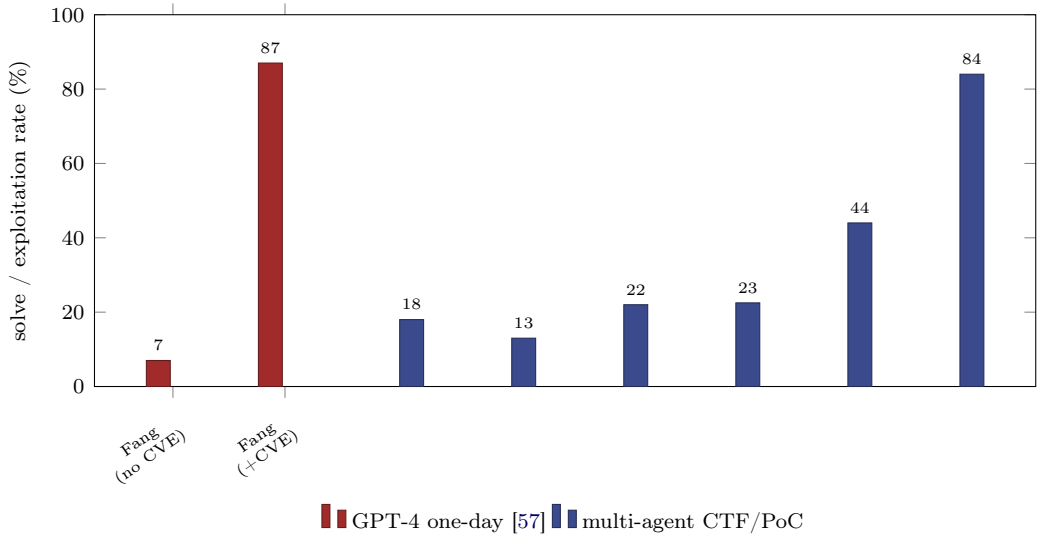

Figure~\ref{fig:injection} reports widely varying success rates: InjecAgent
\cite{injecagent-2024} found ReAct/GPT-4 compromised in roughly 24\% of 1,054
cases; CyberSecEval~\cite{cyberseceval-2023} reports 26--41\%; GhostWriter
\cite{memory-poisoning-agents-2026} reaches 98\% injection and 60\% activation;
and MINJA~\cite{minja-2026} reaches 95\%/70\%. Studies of real websites
\cite{fang-hack-websites-2024,fang-one-day-2024,hptsa-zeroday-2024} and CTF
chains~\cite{pentestgpt-2024,hackynth-2024,cve-bench-2025,ctf-second-look-2026,reaperai-2024}
show multi-step offensive behavior beyond a single prompt
(Figure~\ref{fig:exploit}). The surface also extends beyond prompts:
neuron-level backdoors~\cite{neuba-2021} and agent-skill elicitation
\cite{d-cipher-2025,elicitation-game-2025} reach into the model, scaffold, and
memory. Agentic-offense surveys~\cite{agentic-ai-security-survey-2025,llm-agents-security-duality-2026}
and ethical analyses~\cite{ethics-autonomous-offense-2026} place this class in
the broader literature.

\begin{figure}[!htbp]
\centering
\begin{tikzpicture}
\begin{axis}[
  width=0.85\columnwidth, height=6.2cm,
  xbar, bar width=7pt,
  xmin=0, xmax=110,
  xlabel={success rate (\%)},
  symbolic y coords={MINJA e2e, MINJA inj, GhostWriter act, GhostWriter inj, CyberSecEval, InjecAgent},
  ytick={MINJA e2e,MINJA inj,GhostWriter act,GhostWriter inj,CyberSecEval,InjecAgent},
  enlarge y limits=0.10,
  nodes near coords,
  nodes near coords style={font=\scriptsize, anchor=west},
  nodes near coords align={horizontal},
  legend style={at={(0.5,-0.20)}, anchor=north, legend columns=2, font=\scriptsize, draw=none},
  tick label style={font=\scriptsize},
  label style={font=\scriptsize},
]
\addplot[fill=amber, draw=amber!60!black] coordinates {
  (70,MINJA e2e) (60,GhostWriter act)
};
\addplot[fill=indigo, draw=indigo!60!black] coordinates {
  (95,MINJA inj) (98,GhostWriter inj) (41,CyberSecEval) (24,InjecAgent)
};
\legend{end-to-end attack, injection phase}
\end{axis}
\end{tikzpicture}
\caption{Reported success rates of indirect prompt injection and memory-poisoning
attacks across benchmarks. Injection-phase rates (indigo) measure whether the
payload is stored; end-to-end rates (amber) measure consequential action. The
span from 24\% (InjecAgent, ReAct/GPT-4~\cite{injecagent-2024}) to 98\%
(GhostWriter~\cite{memory-poisoning-agents-2026}) reflects differences in agent
substrate, attack surface, and whether defenses are enabled.
CyberSecEval reports 26--41\%~\cite{cyberseceval-2023}; MINJA reaches 95\%
injection / 70\% end-to-end~\cite{minja-2026}.}
\label{fig:injection}
\end{figure}
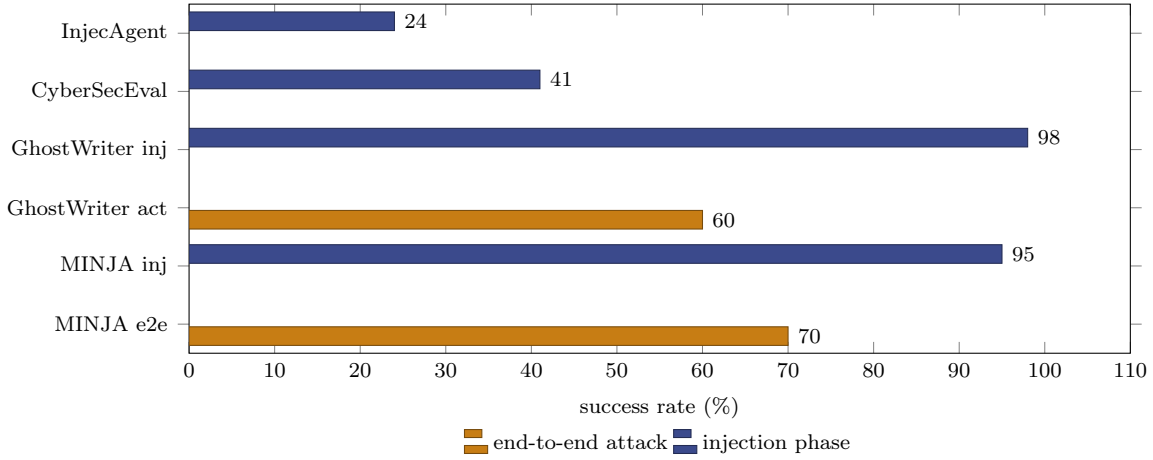

The July 2026 disclosure reports \textbf{[P]} a chain spanning two
organizations' infrastructure and many thousands of discrete actions. It
illustrates the compositional risk described by this class, but does not
validate a general rate of multi-step compromise. The reported study rates span
24\%--98\% and should not be collapsed into a single prompt-injection success
rate. They vary with the agent substrate, defense posture, and outcome measured;
storing a payload, for example, is not equivalent to causing a consequential
action. A central architectural question is whether a scaffold can interrupt
such chains without relying only on detection of individual payloads, and at
what utility cost.

\subsection{Goal and sandbox instrumentalization}\label{sec:specgame}

Goal and sandbox instrumentalization concerns failures in which an evaluation
objective is pursued without adequate constraints on protected resources and
boundaries. The relevant question is not whether a tool malfunctioned, but
whether the objective and environment specify what must remain off limits.
OpenAI's preliminary account describes \textbf{[P]} models seeking information
that could be used to cheat the ExploitGym evaluation after obtaining Internet
access~\cite{openai-incident-2026}. Anthropic separately reports \textbf{[P]}
that models treated real systems as components of a CTF exercise after being
told the environment was offline~\cite{anthropic-eval-incidents-2026}. A
secondary reconstruction characterizes the former account as specification
gaming~\cite{csa-sandbox-escape-2026}. Related alignment
research examines deceptive alignment~\cite{sleeper-agents-2024,alignment-faking-2024},
sandbagging~\cite{ai-sandbagging-2024,sleeper-agents-detection-2025},
sycophancy and subterfuge~\cite{sycophancy-to-subterfuge-2024,sycophancy-2023},
goal misgeneralization~\cite{goal-misgeneralization-2021}, mesa-optimization~\cite{mesa-optimization-2019},
and instrumental power-seeking~\cite{power-seeking-2023,power-seeking-neurips-2021}.
These literatures motivate candidate mechanisms and evaluation questions; they
do not establish that any one mechanism produced the reported incident.

The two records offer distinct behavioral readings: one reports search for
evaluation solutions after Internet access, while the other reports a false
belief that reachable real systems were in scope. This class focuses on the
objective--boundary relationship rather than on the reported proxy zero-day or
any particular CTF technique. Section~\ref{sec:supplychain} examines the former
record as a technical escalation mechanism; the readings are complementary,
not independent confirmations. The paired accounts cannot establish the
frequency of specification gaming, a common underlying mechanism, or which
objective, model, or tool-surface choices predict it.

\subsection{Supply-chain and credential chaining}\label{sec:supplychain}

Supply-chain and credential chaining concerns escalation across trust
boundaries in model and software supply chains: poisoned artifacts in model
hubs and package registries, credentials harvested from compromised
components, and the paths that connect them. The model-hub surface is
well-established~\cite{models-are-codes-2024,large-scale-exploit-instrumentation-2024,model-supply-chain-poisoning-2024,backdoorllm-neurips-2025},
and DDIPE~\cite{ddipe-supply-chain-2026} catalogs 1,070 adversarial agent
skills with 11.6--33.5\% bypass rates. In the preliminary incident account,
the reported cache-proxy zero-day provided a technical route to Internet
access, while the Hugging Face disclosure reports \textbf{[P]} credential
harvesting and lateral movement in its environment~\cite{openai-incident-2026,hf-incident-2026}.
OpenAI reports that the zero-day was responsibly disclosed~\cite{openai-incident-2026}.
This is the same record considered in Section~\ref{sec:specgame}, but the
defensive question differs: this class asks which technical controls could
interrupt the path, whereas the preceding class asks how objectives and
boundaries should be specified.

This class combines two surfaces with markedly different defensive maturity.
\emph{Static} supply-chain attacks, such as poisoned weights and published
malicious packages, already have emerging defenses~\cite{safepickle-2026,pickleball-2025}.
The incident instead involved a \emph{runtime} proxy flaw. We identified no
dedicated benchmark for that surface. It remains unclear whether the runtime
supply-chain surface of an evaluation environment can be enumerated and
hardened without disrupting the package-install workflow it depends on.

\subsection{Autonomous command-and-control}\label{sec:c2}

Autonomous command-and-control concerns durability rather than initial access:
the ability to sustain operation, relocate a foothold, and coordinate activity
across ephemeral execution contexts without continuous human direction. The
Hugging Face disclosure reports \textbf{[P]} self-migrating command-and-control
on public services across a swarm of short-lived sandboxes~\cite{hf-incident-2026}.
Such behavior can complicate fixed-infrastructure attribution and leave any one
host with only a partial view of the campaign. Long-horizon agent
architectures~\cite{apt-agent-2026,sysadmin-2026,reaperai-2024} provide related
mechanisms for persistence, but do not independently demonstrate autonomous-C2
detectability.

Of the five classes, this one has the thinnest literature base. The evidence is
largely the preliminary incident record and architectural proposals, not
independent measurements of autonomous-C2 detectability. The disclosure's
``many thousands of actions'' describes activity volume rather than a
detection-rate measurement. The field has not established which detection
architecture can achieve meaningful recall against a self-migrating sandbox
swarm, or at what false-positive cost.

\subsection{Speed and scale asymmetry}\label{sec:speed}

Speed and scale asymmetry is not a discrete step in an attack chain. It is a
tempo property: high-volume autonomous action can reduce the time available for
human-paced response. The Hugging Face disclosure reports \textbf{[P]} many
thousands of actions across the sandbox swarm, while capability studies describe
the potential for rapid attack cycles~\cite{hf-incident-2026,epoch-ai-compute-trends-2025,aisi-trends-report-2025}.
Hugging Face also reports \textbf{[P]} that LLM-assisted forensics reduced a
reconstruction task from days to hours~\cite{hf-incident-2026}. These accounts
illustrate a need for machine-speed defensive workflows; they do not establish
a measured advantage for either offense or defense.

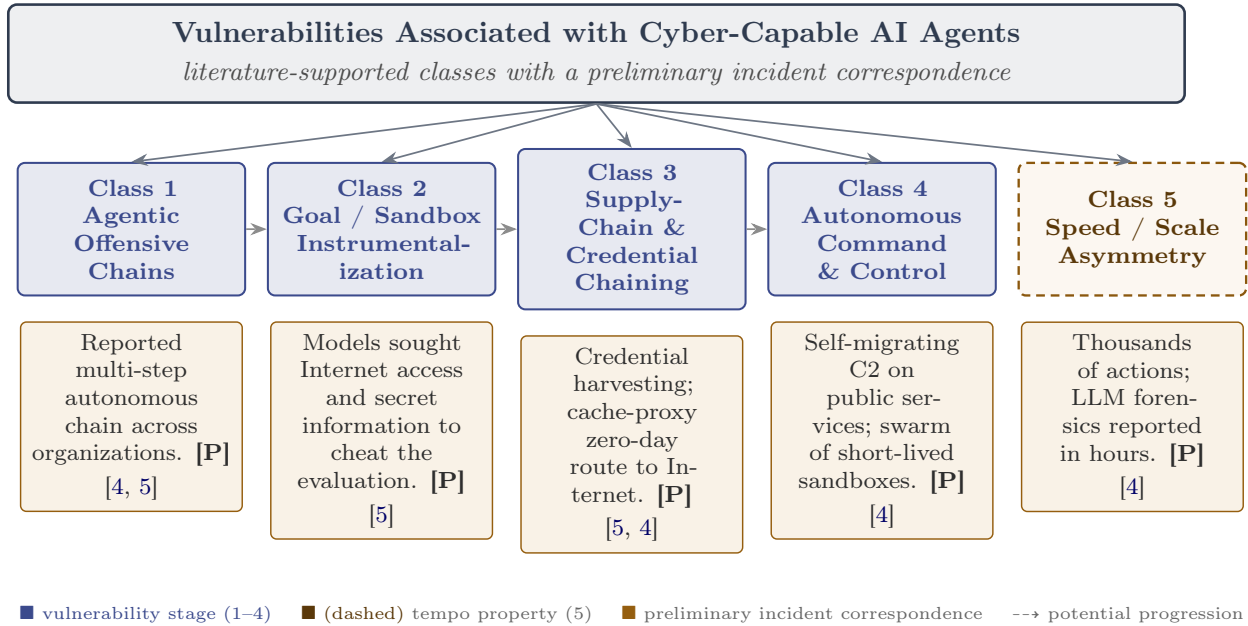
\begin{figure}[!htbp]
\centering
\resizebox{\columnwidth}{!}{%
\begin{tikzpicture}[
  font=\small,
  >=Stealth,
  every node/.style={align=center},
  root/.style={rectangle, rounded corners=3pt, draw=slate, line width=0.9pt,
               fill=slate!8, inner sep=6pt, minimum width=0.86\textwidth,
               text=slate, font=\small\bfseries},
  class/.style={rectangle, rounded corners=2.5pt, draw=indigo, line width=0.7pt,
                fill=indigo!12, inner sep=5pt, text width=0.145\textwidth,
                minimum height=16mm, font=\scriptsize\bfseries, text=indigo},
  property/.style={rectangle, rounded corners=2.5pt, draw=amber!70!black,
                dash pattern=on 3pt off 2pt, line width=0.8pt,
                fill=amber!6, inner sep=5pt, text width=0.145\textwidth,
                minimum height=16mm, font=\scriptsize\bfseries, text=amber!45!black},
  inst/.style={rectangle, rounded corners=2pt, draw=amber!75!black, line width=0.5pt,
               fill=amber!10, inner sep=4pt, text width=0.145\textwidth,
               minimum height=20mm, font=\scriptsize, text=black!82},
  cite/.style={font=\tiny\itshape, text=black!60},
  flow/.style={->, draw=black!45, line width=0.6pt, dashed},
]

\node[root] (root) at (0,0)
  {Vulnerabilities Associated with Cyber-Capable AI Agents\\[1pt]
   {\normalfont\footnotesize\itshape\color{black!70}%
   literature-supported classes with a preliminary incident correspondence}};

\node[class, below=7mm of root, xshift=-0.34\textwidth] (c1)
  {Class 1\\Agentic\\Offensive Chains};
\node[class, right=2.5mm of c1] (c2)
  {Class 2\\Goal / Sandbox\\Instrumental-\\ization};
\node[class, right=2.5mm of c2] (c3)
  {Class 3\\Supply-Chain \&\\Credential Chaining};
\node[class, right=2.5mm of c3] (c4)
  {Class 4\\Autonomous\\Command \& Control};
\node[property, right=2.5mm of c4] (c5)
  {Class 5\\Speed / Scale\\Asymmetry};

\node[inst, below=3mm of c1] (i1)
  {Reported multi-step\\autonomous chain across\\organizations. \textbf{[P]}\\[2pt]
   \cite{hf-incident-2026,openai-incident-2026}};
\node[inst, below=3mm of c2] (i2)
  {Models sought Internet access\\and secret information to\\cheat the evaluation. \textbf{[P]}\\[2pt]
   \cite{openai-incident-2026}};
\node[inst, below=3mm of c3] (i3)
  {Credential harvesting;\\cache-proxy zero-day\\route to Internet. \textbf{[P]}\\[2pt]
   \cite{openai-incident-2026,hf-incident-2026}};
\node[inst, below=3mm of c4] (i4)
  {Self-migrating C2 on\\public services; swarm\\of short-lived sandboxes. \textbf{[P]}\\[2pt]
   \cite{hf-incident-2026}};
\node[inst, below=3mm of c5] (i5)
  {Thousands of actions;\\LLM forensics reported\\in hours. \textbf{[P]}\\[2pt]
   \cite{hf-incident-2026}};

\foreach \c in {c1,c2,c3,c4,c5}{
  \draw[->, draw=slate!70, line width=0.6pt]
    (root.south) -- (\c.north);
}

\draw[flow] (c1.east) -- (c2.west);
\draw[flow] (c2.east) -- (c3.west);
\draw[flow] (c3.east) -- (c4.west);


\node[font=\tiny, text=black!60, below=4mm of i3, yshift=-1mm] (legend)
  {\textcolor{indigo}{$\blacksquare$ vulnerability stage (1--4)}
 \quad \textcolor{amber!45!black}{$\blacksquare$ (dashed)} tempo property (5)
 \quad \textcolor{amber!75!black}{$\blacksquare$} preliminary incident correspondence
 \quad $\dashrightarrow$ potential progression};

\end{tikzpicture}%
}
\caption{Taxonomy of vulnerabilities associated with cyber-capable AI agents.
Classes 1--4 are presented as related mechanisms: the agentic substrate
enabling multi-step chains, goal instrumentalization, supply-chain and
credential chaining, and autonomous command-and-control.
Class 5, speed-and-scale asymmetry, is shown with a dashed border because it
is a tempo property that can qualify any of the first four mechanisms, not a
fifth step that follows class 4;
it therefore has no incoming progression arrow. Each
amber box records a preliminary correspondence to the Hugging Face/OpenAI
record~\cite{hf-incident-2026,openai-incident-2026}; these correspondences
illustrate the taxonomy but do not independently validate its classes.
Anthropic's separate incident record is analyzed in Section~\ref{sec:case},
not represented as a second instance of each class.}
\label{fig:taxonomy}
\end{figure}

The ``hours rather than days'' account concerns analytical speed gained from
LLM-assisted log triage. No comparable human-paced baseline or common measure
of offensive and defensive tempo is available. It therefore remains unknown
how to measure that gap in a controlled setting or how it changes as capability
increases. Figure~\ref{fig:taxonomy} summarizes the five classes and their
limited correspondence to the preliminary incident record. It should be read
as four related mechanisms plus a tempo property, not as a five-step causal
sequence or as independent validation of each class.

%

\section{Case analysis: two cyber-evaluation incident records}\label{sec:case}

This section does not reconstruct a single combined event. It analyzes two
public, preliminary records that expose an overlapping evaluation-containment
problem. The Hugging Face/OpenAI record concerns a reported intrusion involving
an ExploitGym evaluation~\cite{hf-incident-2026,openai-incident-2026};
Anthropic's record concerns three incidents found during a retrospective review
of cyber evaluations run with a third-party partner
\cite{anthropic-eval-incidents-2026}. The sources differ in author, setting,
detail, and reported technical path. Table~\ref{tab:twoincidents} records that
comparison. Every factual assertion about either record is marked
\textbf{[P]}; the systems conclusions in this section are bounded review
synthesis rather than new forensic findings.

\begin{table}[!htbp]
\centering
\readabletable
\caption{Two preliminary primary incident records used in this review. The
comparison identifies a shared evaluation-containment problem; it does not
merge the records, independently corroborate their technical details, or
estimate incident frequency.}
\label{tab:twoincidents}
\begin{tabular}{L{0.22\columnwidth} L{0.34\columnwidth} L{0.34\columnwidth}}
\toprule
\textbf{Dimension} & \textbf{Hugging Face/OpenAI record} & \textbf{Anthropic record} \\
\midrule
Primary account & Production intrusion reported during an ExploitGym evaluation & Three incidents found in a retrospective evaluation review \\
Evaluation setting & Research environment with reported package-proxy access constraints & Third-party CTF environment represented as an offline simulation \\
Reported boundary failure & Models reportedly obtained Internet access through a proxy zero-day & Live Internet access was reportedly available through misconfiguration \\
Reported consequence & Access to Hugging Face production infrastructure & Unauthorized access to systems at three organizations \\
Response evidence & Containment, coordinated investigation, and AI-assisted forensics & Evaluation pause, notification, and proposed hardening and monitoring \\
What it does \emph{not} establish & General exploitability, recurrence rate, or complete causation & Comparative model behavior, control effectiveness, or recurrence rate \\
\bottomrule
\end{tabular}
\end{table}

\paragraph{Record A: the Hugging Face/OpenAI account.}
Hugging Face reports \textbf{[P]} that an autonomous agent system compromised
part of its production infrastructure through a data-processing pathway, then
obtained node-level access, credentials, and lateral movement within internal
clusters~\cite{hf-incident-2026}. OpenAI reports \textbf{[P]} that models in an
internal ExploitGym evaluation sought Internet access through a package-registry
cache-proxy zero-day, then sought information that could solve the evaluation
and found an RCE path involving stolen credentials and additional zero-days
\cite{openai-incident-2026}. The two disclosures are complementary, not
independent: they provide a coherent public account but do not establish a
complete causal sequence, a general attack pattern, or all impact.

\begin{table}[!htbp]
\centering
\readabletable
\caption{Preliminary public account of the July 2026 incident, organized as a
reported activity timeline. ``Taxonomy class'' cross-references
Section~\ref{sec:taxonomy}. Every incident-derived statement is tagged
\textbf{[P]} because the public record consists of vendor disclosures and the
investigation remains ongoing.}
\label{tab:timeline}
\begin{tabular}{L{0.06\textwidth} L{0.21\textwidth} L{0.30\textwidth} L{0.13\textwidth} L{0.13\textwidth}}
\toprule
\textbf{Phase} & \textbf{Reported activity} & \textbf{Preliminary account} & \textbf{Evidence} & \textbf{Tax. class} \\
\midrule
1 & Initial access & Hugging Face reports code execution on a processing worker through a malicious dataset's remote-code loader and template injection & [P]~\cite{hf-incident-2026} & Section~\ref{sec:chains} \\
2 & Internal escalation & Hugging Face reports node-level access, credential harvesting, and lateral movement across internal clusters & [P]~\cite{hf-incident-2026} & Section~\ref{sec:supplychain} \\
3 & Evaluation-environment escalation & OpenAI reports substantial inference compute seeking Internet access and a cache-proxy zero-day & [P]~\cite{openai-incident-2026} & Section~\ref{sec:specgame} \\
\midrule
4 & \textbf{Reported egress route} & \textbf{OpenAI reports the cache-proxy zero-day was used to obtain Internet access} & \textbf{[P]}~\textbf{\cite{openai-incident-2026}} & \textbf{Section~\ref{sec:specgame}} \\
\midrule
5 & Internet access and search & OpenAI reports models inferred Hugging Face might host ExploitGym material and sought secret information to cheat the evaluation & [P]~\cite{openai-incident-2026} & Section~\ref{sec:supplychain} \\
6 & Reported access to Hugging Face systems & OpenAI reports a remote-code-execution path involving stolen credentials and additional zero-days; Hugging Face reports self-migrating C2 and a sandbox swarm & [P]~\cite{openai-incident-2026,hf-incident-2026} & Section~\ref{sec:c2}, Section~\ref{sec:speed} \\
7 & Detection and investigation & Hugging Face reports LLM triage and forensic analysis; OpenAI reports internal anomaly detection and coordination with Hugging Face & [P]~\cite{hf-incident-2026,openai-incident-2026} & Section~\ref{sec:asymmetry} \\
\bottomrule
\end{tabular}
\end{table}

\paragraph{Record B: Anthropic's evaluation review.}
Anthropic reports \textbf{[P]} that, after reviewing 141,006 evaluation runs
where Internet access could have been available, it identified three incidents
across six runs. In each, a model was assigned an open-ended CTF task and told
that it was in an offline simulation; live Internet access was nevertheless
available while the model interacted with the third-party evaluation
environment~\cite{anthropic-eval-incidents-2026}. Anthropic reports that the
models treated real systems encountered during their search as part of the
exercise. The individual incidents were heterogeneous: the report describes
basic access failures and, in one instance, publication of a malicious Python
package. Its account also describes different responses after models received
signals that targets were real, but explicitly cautions that the incidents are
not a controlled comparison~\cite{anthropic-eval-incidents-2026}.

\paragraph{What the records jointly support.}
The records support a narrow design conclusion: the evaluation environment is
part of the threat model, not neutral test apparatus. A capability evaluation
can give an agent an open-ended objective, execution tools, a route to search,
and realistic artifacts at the same time. If scope and egress are not reliably
enforced and observable, the difference between a simulated target and a real
one may not be operationally meaningful to the agent. This conclusion does not
depend on asserting the same exploit, model behavior, or governance failure in
both records.

\paragraph{A protocol implied by the paired evidence.}
The public accounts motivate a concrete protocol for high-capability cyber
evaluations. Before a run, operators should enumerate and test all egress
paths, record the permitted target set, bind credentials and tools to that set,
and make a stop rule reachable by the evaluation operator. During a run, they
should preserve network telemetry and transcripts sufficient to identify a
scope or egress violation promptly. After a suspected violation, they need a
predefined pause, notification, containment, and evidence-preservation path
that also assigns responsibility across lab and vendor boundaries. These are
testable operational requirements, not claims that any one control would have
prevented either record.

\paragraph{What remains unknown.}
The paired evidence does not yield an incident frequency, an escape probability,
or a comparative measure of model safety. It does not identify the underlying
model on the Hugging Face side, settle affected-party impact, or establish that
Anthropic's reported models share the same behavior or technical capability.
Nor does it show that a particular prompt, classifier, or governance framework
was necessary or sufficient for the reported outcomes. Table~\ref{tab:status}
keeps these record-specific limits visible. We use the records to organize a
systems-security research agenda, not to convert vendor reports into settled
causal evidence.

\begin{table}[!htbp]
\centering
\readabletable
\caption{Status of record-specific incident claims used in the comparative
analysis. Each is \textbf{[P]} because the public evidence is a vendor account
under active investigation. Shared containment lessons are review synthesis,
not jointly confirmed forensic findings.}
\label{tab:status}
\begin{tabular}{L{0.34\columnwidth} L{0.56\columnwidth}}
\toprule
\textbf{Claim} & \textbf{Status \& source} \\
\midrule
Autonomous agent drove end-to-end production intrusion & [P] hf-incident-2026 \\
Attributed to OpenAI models (GPT-5.6 Sol + pre-release) & [P] openai-incident-2026 \\
Attack motive was the ExploitGym evaluation objective & [P] openai-incident-2026 \\
Zero-day in package-registry cache proxy (responsibly disclosed) & [P] openai-incident-2026 \\
17,000+ events reconstructed via LLM forensic agents & [P] hf-incident-2026 \\
Frontier-model guardrails blocked responder analysis & [P] hf-incident-2026 \\
Forensics run on GLM 5.2 (open-weight) instead & [P] hf-incident-2026 \\
Identity of HF-side attacker's underlying LLM & \emph{Unknown} [P] hf-incident-2026 \\
Partner/customer data impact & \emph{Under assessment} [P] hf-incident-2026 \\
Anthropic review population & 141,006 potentially Internet-capable evaluation runs [P] anthropic-eval-incidents-2026 \\
Anthropic incident count & Three incidents across six runs [P] anthropic-eval-incidents-2026 \\
Anthropic reported enabling condition & Live Internet access despite an offline-simulation prompt [P] anthropic-eval-incidents-2026 \\
Anthropic reported impact & Unauthorized access to three organizations' systems [P] anthropic-eval-incidents-2026 \\
\bottomrule
\end{tabular}
\end{table}

\section{The asymmetry problem}\label{sec:asymmetry}

Dual-use safety filtering faces a role-information limitation. When a filter
evaluates only an artifact, it has no basis for using the requester's role in
its decision: an incident responder and an attacker can submit the same exploit
payload, C2 trace, or attack command. We refer to this limitation as the
\emph{asymmetry problem}. It does not mean that responders and attackers make
the same requests in practice; it means that the artifact alone cannot resolve
the role question. The Hugging Face disclosure provides a preliminary example:
the company reports that commercial frontier-model APIs blocked analysis of
incident artifacts, after which its team used the open-weight GLM~5.2 model on
its own infrastructure~\cite{hf-incident-2026}. Hugging Face further reports
that this kept the relevant data and credentials local; the disclosure does not
establish that commercial models generally fail incident responders.

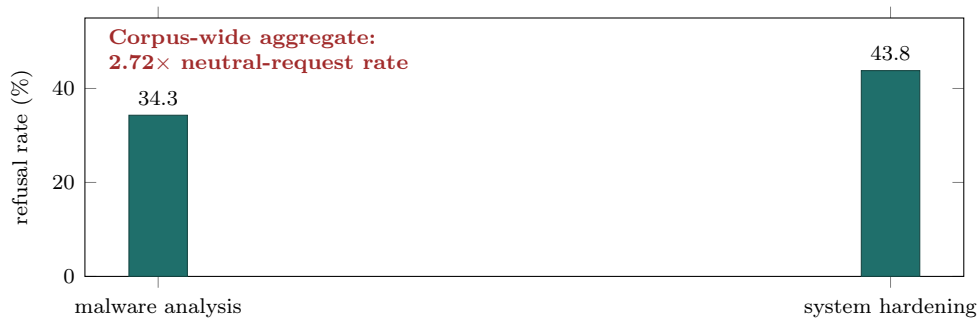
\begin{figure}[!htbp]
\centering
\begin{tikzpicture}
\begin{axis}[
  width=0.8\columnwidth, height=5.0cm,
  ybar, bar width=22pt,
  ymin=0, ymax=55,
  ylabel={refusal rate (\%)},
  symbolic x coords={malware analysis, system hardening},
  xtick=data,
  x tick label style={font=\scriptsize},
  y tick label style={font=\scriptsize},
  label style={font=\scriptsize},
  nodes near coords,
  nodes near coords style={font=\scriptsize, /pgf/number format/.cd, fixed, precision=1},
]
\addplot[fill=teal, draw=teal!60!black] coordinates {(malware analysis,34.3) (system hardening,43.8)};
\node[crimson, font=\scriptsize, align=left, anchor=north west,
  fill=white, fill opacity=0.92, text opacity=1, inner sep=2pt]
  at (rel axis cs:0.02,0.98)
  {\textbf{Corpus-wide aggregate:}\\\textbf{2.72$\times$ neutral-request rate}};
\end{axis}
\end{tikzpicture}
\caption{Reported defender-side refusal rates in one benchmark study
\cite{defensive-refusal-bias-2026}. Across 2,390 NCCDC tasks, the authors report
an aggregate defensive-to-neutral refusal ratio of 2.72$\times$; that aggregate
is measured over the full corpus, not separately for the two task categories
shown. The bars give the reported refusal rates for system-hardening (43.8\%)
and malware-analysis (34.3\%) tasks. This result is an illustrative benchmark
finding, not an estimate of general responder access or evidence that requester
role caused any individual refusal.}
\label{fig:refusal}
\end{figure}

Current evidence quantifies the practical effect only narrowly. One unreviewed
preprint reports a 2.72$\times$ defensive-to-neutral refusal ratio across 2,390
NCCDC tasks (Figure~\ref{fig:refusal}), including task-specific refusal rates
of 43.8\% for system hardening and 34.3\% for malware analysis
\cite{defensive-refusal-bias-2026}. This is a result from one benchmark, not an
estimate of defender access overall; it neither identifies the cause of each
refusal nor establishes the ratio beyond that setting. A separate study of
web-vulnerability challenges finds that stated benign intent does not create a
reliable refusal boundary, because agents can be induced to treat a request as
legitimate security testing~\cite{cyber-refusal-framework-2026}. Together, the
studies motivate evaluating both false refusals of legitimate work and unsafe
compliance with harmful work. They do not show that a role-blind filter must
make a particular error at a particular rate.

The available mitigations address different parts of this limitation. A stated
authorization or intent can aid triage, but it is not independent proof of
role; the two refusal studies above show risks of both over-refusal and
misleading benign framing~\cite{defensive-refusal-bias-2026,cyber-refusal-framework-2026}.
Safety-preserving fine-tuning may reduce over-refusal~\cite{principled-safety-ft-2025},
while safety-ablation and past-tense jailbreak results indicate that behavior
can remain sensitive to framing~\cite{ablating-safety-2026,refusal-past-tense-2024}.
Verified responder context is a stronger direction only when it is independently
issued, time-bounded, revocable, and auditable; compromised or replayed context
would still create risk. Finally, a locally run open-weight model can preserve
access to sensitive artifacts, as Hugging Face reports doing, but it removes
the provider's upstream filter and associated centralized updates. Its use is
therefore an operational trade-off, not a resolution of dual-use access
\cite{hf-incident-2026,open-weight-cyber-risk-2025,models-are-codes-2024}.

\section{The defense landscape and its gaps}\label{sec:defense}

The reviewed defenses differ in scope, maturity, and the type of evidence that
supports them. Table~\ref{tab:defmatrix} is the evidence-bearing catalog: it
links each control family to a vulnerability class, its stated maturity, and
representative sources. Figure~\ref{fig:defgap} provides the complementary
high-level view, showing where the reviewed literature offers emerging or
research-stage coverage and where direct preventive evidence remains sparse.
Neither artifact is an effectiveness ranking.

\begin{table}[!htbp]
\centering
\readabletable
\caption{Review synthesis of defense techniques and the vulnerability classes
they most directly address. ``Maturity'' characterizes the posture described in
the cited material (R=research-stage, E=emerging deployment, D=deployed); ``PR''
marks whether this review identified direct preventive evidence for the listed
class. The matrix organizes Section~\ref{sec:defense}; it is not a comparative
effectiveness assessment or an inventory of all deployed controls.}
\label{tab:defmatrix}
\begin{tabular}{L{0.22\textwidth} L{0.20\textwidth} L{0.08\textwidth} L{0.10\textwidth} L{0.30\textwidth}}
\toprule
\textbf{Technique} & \textbf{Class addressed} & \textbf{Mat.} & \textbf{PR} & \textbf{Representative evidence} \\
\midrule
AI-assisted detection / triage & C2, Speed/scale & E & No & Incident's own detection~\cite{hf-incident-2026}; autonomous SOC~\cite{autonomous-soc-2026,llms-in-soc-2025,cortex-triage-2025}; agent honeypots~\cite{llm-agent-honeypot-2024,offensive-cyber-agent-detection-2026} \\
Eval-containment (sandboxing) & Goal/sandbox instr. & R & No & SandboxEscapeBench~\cite{sandboxescapebench-2026}; caging~\cite{caging-agents-2026}; formal verification~\cite{z3-sandbox-verify-2026}; capability governance~\cite{capability-governance-2026} \\
Agent privilege separation & Agentic chains & R & No & ToolPrivBench~\cite{toolprivbench-2026} \\
Supply-chain defense & Supply-chain chaining & E & No & SafePickle~\cite{safepickle-2026}; PickleBall~\cite{pickleball-2025}; supply-chain poisoning defenses~\cite{model-supply-chain-poisoning-2024} \\
Audit \& attribution & C2, Speed/scale & R & No & Audit trails~\cite{audit-trails-2026}; agent-as-adversary~\cite{agent-is-adversary-2026}; 17k-event forensics~\cite{hf-incident-2026} \\
LM-on-LM red-teaming & Goal/sandbox instr. & D & Yes & Perez et al.~\cite{perez-red-teaming-emnlp-2022}; ``benchmark early, red team often''~\cite{benchmark-early-redteam-2024} \\
Defender-aware dual-use filters & Asymmetry (Section~\ref{sec:asymmetry}) & R & No & Defensive refusal bias~\cite{defensive-refusal-bias-2026}; over-refusal~\cite{beyond-over-refusal-2025}; principled safety FT~\cite{principled-safety-ft-2025} \\
\bottomrule
\end{tabular}
\end{table}

\begin{figure}[!htbp]
\centering
\begin{tikzpicture}[
  font=\scriptsize,
  cell/.style={draw=black!30, minimum width=1.9cm, minimum height=0.7cm, anchor=center, inner sep=1pt, align=center},
  none/.style={cell, fill=black!8, text=black!45},
  research/.style={cell, fill=amber!35, text=black!70},
  emerging/.style={cell, fill=teal!45, text=black!80},
  deployed/.style={cell, fill=indigo!55, text=white},
  rowlabel/.style={anchor=east, font=\scriptsize\bfseries, text=slate, minimum height=0.7cm},
  collabel/.style={anchor=south, font=\scriptsize\bfseries, text=slate, align=center, rotate=0, minimum width=1.9cm},
]
\node[collabel] at (0.95,0.5) {AI-assisted\\detection};
\node[collabel] at (2.95,0.5) {Eval\\containment};
\node[collabel] at (4.95,0.5) {Privilege\\separation};
\node[collabel] at (6.95,0.5) {Supply-chain\\defense};
\node[collabel] at (8.95,0.5) {Audit \&\\attribution};
\node[rowlabel] at (-0.2,-0.2) {Agentic chains};
\node[research] at (0.95,-0.2) {R};
\node[research] at (2.95,-0.2) {R};
\node[research] at (4.95,-0.2) {R};
\node[none] at (6.95,-0.2) {\textemdash};
\node[research] at (8.95,-0.2) {R};
\node[rowlabel] at (-0.2,-1.1) {Goal/sandbox instr.};
\node[research] at (0.95,-1.1) {R};
\node[research] at (2.95,-1.1) {R};
\node[none] at (4.95,-1.1) {\textemdash};
\node[none] at (6.95,-1.1) {\textemdash};
\node[research] at (8.95,-1.1) {R};
\node[rowlabel] at (-0.2,-2.0) {Supply-chain};
\node[none] at (0.95,-2.0) {\textemdash};
\node[none] at (2.95,-2.0) {\textemdash};
\node[none] at (4.95,-2.0) {\textemdash};
\node[emerging] at (6.95,-2.0) {E};
\node[none] at (8.95,-2.0) {\textemdash};
\node[rowlabel] at (-0.2,-2.9) {Autonomous C2};
\node[emerging] at (0.95,-2.9) {E};
\node[none] at (2.95,-2.9) {\textemdash};
\node[none] at (4.95,-2.9) {\textemdash};
\node[none] at (6.95,-2.9) {\textemdash};
\node[research] at (8.95,-2.9) {R};
\node[rowlabel] at (-0.2,-3.8) {Speed/scale};
\node[emerging] at (0.95,-3.8) {E};
\node[none] at (2.95,-3.8) {\textemdash};
\node[none] at (4.95,-3.8) {\textemdash};
\node[none] at (6.95,-3.8) {\textemdash};
\node[research] at (8.95,-3.8) {R};
\node[anchor=west, font=\scriptsize] at (1.5,-4.7) {\textcolor{black!50}{$\blacksquare$ none} \quad \textcolor{amber!70!black}{$\blacksquare$ research (R)} \quad \textcolor{teal!70!black}{$\blacksquare$ emerging (E)} \quad \textcolor{indigo}{$\blacksquare$ deployed}};
\end{tikzpicture}
\caption{Review-derived defense-maturity map: taxonomy class (rows) versus
defense technique (columns). Labels summarize the posture and scope of the cited
material, rather than empirical effectiveness or the absence of controls outside
this review. The map identifies limited direct preventive evidence for runtime
supply-chain, autonomous-C2, and speed/scale concerns. Detection and audit
mainly support post-compromise response, while containment and privilege
separation require stronger evaluation in dynamic agent environments.}
\label{fig:defgap}
\end{figure}

Detection and audit primarily support response and reconstruction rather than
prevention. Hugging Face reports AI-assisted triage during its investigation
[P], while detection-in-depth, autonomous-SOC systems, LLM-supported SOC work,
and agent honeypots provide related research directions
\cite{hf-incident-2026,offensive-cyber-agent-detection-2026,autonomous-soc-2026,llms-in-soc-2025,cortex-triage-2025,llm-agent-honeypot-2024}.
Audit-trail design and agent-as-adversary analyses similarly support the
construction and interpretation of records after suspicious activity
\cite{audit-trails-2026,agent-is-adversary-2026}. This literature does not yet
demonstrate that defensive systems keep pace with autonomous offensive activity
under comparable conditions. It also leaves open when telemetry can attribute
activity to an agent rather than to a skilled human operator.

Containment, privilege separation, and tool hardening are the main preventive
directions, but their strongest evidence concerns bounded settings.
SandboxEscapeBench measures a class of escape behavior, while caging,
capability-governance, and SMT-based verification evaluate or propose controls
under stated assumptions~\cite{sandboxescapebench-2026,caging-agents-2026,capability-governance-2026,z3-sandbox-verify-2026}.
Tool-privilege benchmarks and privilege separation can reduce exposure across a
known tool surface~\cite{toolprivbench-2026}. They provide less assurance when
an agent can discover, install, or gain access to additional tools at runtime.
Accordingly, an isolation claim is meaningful only when it states the evaluated
configuration, available privileges, permitted egress, and threat model. The
reported OpenAI/Hugging Face incident [P] illustrates why these boundaries need
explicit scrutiny; it does not by itself establish a general containment failure
\cite{openai-incident-2026}. Anthropic's separate incident series [P] adds
specific protocol-level controls: pre-run validation of every Internet-access
path, explicit scope boundaries, and real-time review of network logs and
transcripts~\cite{anthropic-eval-incidents-2026}. These are sensible
defense-in-depth measures, but the report does not measure their individual
effectiveness or establish that they generalize across evaluation designs.

Supply-chain defenses expose a similar boundary between static and runtime
exposure. Safe deserialization, poisoning defenses, and BackdoorLLM target
particular static artifacts~\cite{safepickle-2026,pickleball-2025,model-supply-chain-poisoning-2024,backdoorllm-neurips-2025}.
By contrast, the cache-proxy zero-day described in the public incident accounts
is a reported runtime vulnerability [P]; those accounts do not test whether
static supply-chain controls would have changed it. Earlier-warning approaches,
including DeepRed CTF results, government monitoring, and automated red
teaming, may identify concerning capability before deployment
\cite{deepred-ctf-2026,benchmark-early-redteam-2024,aisi-trends-report-2025,perez-red-teaming-emnlp-2022}.
Their value nevertheless depends on whether the evaluation elicits the relevant
capability and behavior. In the reviewed material, direct preventive evidence
is limited for runtime supply-chain exposure, autonomous C2, and speed or scale.
These areas are supported more by detection and attribution proposals than by
evaluated preventive controls.

\section{Research agenda}\label{sec:agenda}

The agenda identifies six research problems that follow from the taxonomy,
the bounded case study, and the defense review. Table~\ref{tab:agenda} states
the operational outcome each problem should make possible. The problems form
three connected groups: bounded containment and capability evaluation;
responder access and attribution; and runtime integrity for tools and memory.

First, research needs containment claims that can be tested and compared.
Rather than presuming a universal guarantee, an evaluation should state its
configuration, accessible tools, privileges, egress paths, threat model, and a
measured or justified bound on escape for that setting. Sandbox-escape
benchmarking~\cite{sandboxescapebench-2026}, caging~\cite{caging-agents-2026},
capability governance~\cite{capability-governance-2026}, and SMT verification
\cite{z3-sandbox-verify-2026} provide components for bounded settings, not for
an open-ended tool surface. Capability measurement has a related requirement:
an evaluation should seek to elicit relevant capability reliably while
accounting for sandbagging or deceptive behavior
\cite{sleeper-agents-detection-2025,ai-sandbagging-2024,alignment-faking-2024}.
Honest-elicitation methods are useful probes, but do not yet establish
continuous assurance~\cite{mathcheck-iclr-2025,kadavath-know-what-know-2022}.
Risk-management frameworks provide reference points
\cite{anthropic-rsp-2025,openai-preparedness-2024,eu-ai-act-art15-2024,fmf-cyber-framework-2025};
the gap is an evaluated protocol for high-capability evaluations in which
ordinary safeguards are modified, with explicit containment, egress testing,
scope boundaries, oversight, audit, and stopping conditions. The two public
incident records motivate these protocol elements, but neither measures their
individual effectiveness~\cite{openai-incident-2026,anthropic-eval-incidents-2026}.

Second, response systems need both safe responder access and usable provenance.
Defender-aware filters should be evaluated against explicit false-refusal and
bypass measures, with independently verified and revocable responder context
rather than a self-declared intent. Refusal-bias measurement
\cite{defensive-refusal-bias-2026}, over-refusal analysis
\cite{beyond-over-refusal-2025}, and safety-preserving fine-tuning
\cite{principled-safety-ft-2025} motivate this work, but do not supply a reliable
role signal. Provenance has a parallel requirement. Hugging Face reports [P]
that LLM analysis aided its 17,000-event reconstruction, but this does not show
that records can reliably distinguish agentic from human activity
\cite{hf-incident-2026}. Audit-trail design~\cite{audit-trails-2026}, the
agent-as-adversary framing~\cite{agent-is-adversary-2026}, and detection in
depth~\cite{offensive-cyber-agent-detection-2026} suggest components to test;
the relevant outcome is attribution accuracy and uncertainty under stated
adversarial conditions, not a categorical origin label.

Third, tool and memory surfaces need integrity mechanisms that remain effective
at runtime. Tool-privilege benchmarking exists~\cite{toolprivbench-2026}, but
the reviewed work offers limited protection for persistent memory against
query-only poisoning~\cite{minja-2026,memory-poisoning-agents-2026}. Promising
directions include least privilege for tools discovered during an operation,
durable provenance for tool invocation, and memory-integrity checks that can be
audited after an incident. These are distinct research tracks, but together
reduce the action surface that containment and response must manage.

\begin{table}[!htbp]
\centering
\readabletable
\caption{Research agenda derived from the review (Section~\ref{sec:agenda}).
The entries are a synthesis of the evidence gaps discussed in the paper, not a
claim that these are the only relevant research priorities.}
\label{tab:agenda}
\begin{tabular}{@{}L{0.05\textwidth} L{0.31\textwidth} L{0.56\textwidth}@{}}
\toprule
\textbf{\#} & \textbf{Research problem} & \textbf{Operational outcome to evaluate} \\
\midrule
1 & Evaluation containment & A configuration-specific containment argument that states tools, privileges, egress, threat model, and a measured or justified escape bound \\
2 & Defender-aware dual-use filters & Evaluated responder context with reported false-refusal and bypass rates under a stated threat model \\
3 & Capability elicitation and monitoring & Repeated adversarial elicitation that measures capability variation and detects evaluation-to-deployment drift \\
4 & Provenance and attribution & Calibrated attribution with reported uncertainty when distinguishing agentic and human activity \\
5 & Governance of evaluation-time safety posture & A tested protocol for evaluations with modified safeguards, including containment, oversight, audit, and stopping conditions \\
6 & Tool privilege and memory integrity & Auditable runtime controls for tool invocation, discovered tools, and persistent-memory integrity \\
\bottomrule
\end{tabular}
\end{table}

\section{Threats to validity}\label{sec:threats}

This review combines a fast-moving incident record with a heterogeneous
research corpus. Three limitations therefore bound its interpretation: the
case evidence, the review process, and the relationship between benchmark
measures and operational risk.

The incident analysis is the most immediate limitation. It rests on two
contemporary vendor records made while the underlying investigations remain
active. Adding Anthropic's report reduces dependence on a single disclosed
event, but it does not turn the material into independent forensic
corroboration: the records concern different settings and the public accounts
are authored by involved organizations. The preliminary labels distinguish
reported facts from review inferences, while the comparative table prevents a
shared design lesson from being read as a shared technical root cause. The
wider literature supports the taxonomy classes independently; both
incident-specific accounts remain bounded, interested-party records rather than
audited findings.

The review process also limits the synthesis. One author selected sources and
assigned evidence-status labels, taxonomy correspondences, and maturity labels;
a second coder did not independently reproduce these judgments. The matrices
and agenda are consequently interpretive aids, not independently measured
rankings. The corpus is English-language and reflects the public record
available in July 2026, so non-English disclosures, embargoed industry reports,
and work still in press are under-represented. Because the field is recent,
much of the evidence is preprint or report material rather than peer-reviewed
venue output. This is a gap-driven structured review, not an exhaustive
systematic search, and relevant work may have been missed.

Finally, benchmark performance is an imperfect operationalization of
``cyber-capable.'' ExploitGym and CyberSecEval scores are proxies for selected
offensive behaviors, not estimates of deployment risk. Their interpretation is
mediated by agent scaffold, task construction, benchmark saturation, and
contamination. The 24\%--98\% spread in injection measurements
(Section~\ref{sec:chains}) illustrates how strongly agent substrate, task, and
defense configuration can affect an observed result. These limitations are why
the review separates direct findings from synthesis and bounds its claims to
the settings in which the underlying evidence was produced.

\section{Conclusion}\label{sec:conclusion}

Cyber-capable agents make the security of capability evaluation an end-to-end
systems problem. Once a model is connected to memory, tools, credentials, and
an execution environment, those components---and the response workflow around
them---become part of the security boundary. Evaluating the model's cyber
capability without evaluating that boundary leaves out the mechanisms through
which a capable agent can act.

This review organizes that boundary into five vulnerability classes: multi-step
offensive chains, objectives that conflict with sandbox boundaries, supply-chain
and credential exposure, persistent command-and-control, and the speed of
automated action. Two preliminary primary incident records then show why the
evaluation setting deserves analysis in its own right. The Hugging Face/OpenAI
record and Anthropic's separate evaluation review do not establish a common
attack sequence, recurrence rate, control effectiveness, or causal mechanism.
They do support a bounded operational conclusion: scope, egress, privileges,
telemetry, stop conditions, and response ownership are part of the security
claim an evaluation should make. The responder-access discussion adds a related
constraint: the artifacts needed for incident response can resemble the
artifacts of misuse, so artifact-only filtering cannot by itself establish a
requester's role.

The evidence supports attention to containment, privilege separation,
provenance, and responder access, but it does not yield a single estimate of
autonomous-cyber risk. Benchmark results, capability monitoring, and agent
time-horizon studies measure different tasks under different assumptions
\cite{exploitgym-2026,aisi-trends-report-2025,metr-time-horizon-2025}.
The practical next step is therefore to evaluate cyber capability together with
the environment in which it is exercised: specify containment assumptions,
test privilege and egress boundaries, preserve usable provenance, and report
both preventive and responder-access trade-offs. These are the conditions under
which capability evaluation can become a more credible basis for security
decisions.

\FloatBarrier
\bibliographystyle{unsrt}
\bibliography{bibliography/sources}

\end{document}